\documentclass{article} 
\usepackage{iclr2016_conference}
\usepackage{hyperref}
\usepackage{multirow}
\usepackage{url}
\usepackage{amsmath,amsfonts}
\usepackage{graphicx}
\usepackage{subcaption}
\usepackage{xspace}

\title{Diversity Networks: \\Neural Network Compression Using Determinantal Point Processes}

\author{Zelda Mariet \textrm{and} Suvrit Sra \\
Massachusetts Institute of Technology\\
Cambridge, MA 02139, USA \\
\texttt{zelda@csail.mit.edu,suvrit@mit.edu} \\
}

\newcommand{\Yc}{\mathcal{Y}}
\newcommand{\set}[1]{\mathbb{#1}}
\newcommand{\Pc}{\mathcal{P}}
\newcommand{\MNIST}{\texttt{MNIST}}
\newcommand{\ROT}{\texttt{MNIST\_ROT}}
\newcommand{\CIFAR}{\texttt{CIFAR-10}}
\DeclareMathOperator{\Tr}{Tr}
\graphicspath{ {figures/} }

\newcommand{\algo}{\textsc{Divnet}\xspace}

\iclrfinalcopy

\begin{document}

\maketitle

\begin{abstract}
  We introduce \algo, a flexible technique for learning networks with diverse neurons. \algo models neuronal diversity by placing a Determinantal Point Process (DPP) over neurons in a given layer. It uses this DPP to select a subset of diverse neurons and subsequently fuses the redundant neurons into the selected ones. Compared with previous approaches, \algo offers a more principled, flexible technique for capturing neuronal diversity and thus implicitly enforcing regularization. This enables effective auto-tuning of network architecture and leads to smaller network sizes without hurting performance. Moreover, through its focus on diversity and neuron fusing, \algo remains compatible with other procedures that seek to reduce memory footprints of networks. We present experimental results to corroborate our claims: for pruning neural networks, \algo is seen to be notably superior to competing approaches.
\end{abstract}

\section{Introduction}
Training neural networks requires setting several hyper-parameters to adequate values: number of hidden layers, number of neurons per hidden layer, learning rate, momentum, dropout rate, etc. Although tuning such hyper-parameters via parameter search has been recently investigated by~\citet{maclaurin-2015}, doing so remains extremely costly, which makes it imperative to develop more efficient techniques.

Of the many hyper-parameters, those that determine the network's architecture are among the hardest to tune, especially because changing them during training is more difficult than adjusting more dynamic parameters such as the learning rate or momentum. Typically, the architecture parameters are set once and for all before training begins. Thus, assigning them correctly is paramount: if the network is too small, it will not learn well; if it is too large, it may take significantly longer to train while running the risk of overfitting. Networks are therefore usually trained with more parameters than necessary, and pruned once the training is complete.

This paper introduces \algo, a new technique for reducing the size of a network. \algo decreases the amount of redundancy in a neural network (and hence its size) in two steps: first, it samples a diverse subset of neurons; then, it merges the remaining neurons with the ones previously selected. 

Specifically, \algo models neuronal diversity by placing a Determinantal Point Process (DPP)~\citep{hough06} over neurons in a layer, which is then used to select a subset of diverse neurons. Subsequently,  \algo ``fuses''  information from the dropped neurons into the selected ones through a reweighting procedure. Together, these steps reduce network size (and act as implicit regularization), without requiring any further training or significantly hurting performance. \algo is fast and runs in time negligible compared to the network's prior training time. Moreover, it is agnostic to other network parameters such as activation functions, number of hidden layers, and learning rates.

For simplicity, we describe and analyze \algo for feed-forward neural networks, however \algo is \emph{not} limited to this setting. Indeed, since \algo operates on a layer fully connected to the following one in a network's hierarchy, it applies equally well to other architectures with fully connected layers. For example, it can be applied without any further modification to Deep Belief Nets and to the fully-connected layers in Convolutional Neural Networks. As these layers are typically responsible for the large majority of the CNNs' memory footprint~\citep{yang-2014}, \algo is particularly well suited for such networks.

\paragraph{Contributions.} The key contributions of this paper are the following:
\begin{list}{--}{\leftmargin=1em}
  \setlength{\itemsep}{1pt}
\item Introduction of DPPs as a flexible, powerful tool for modeling layerwise neuronal diversity (\S\ref{sec:pruning}). Specifically, we present a practical method for creating DPPs over neurons, which enables diversity promoting sampling and thereby leads to smaller network sizes.
  \item A simple but crucial ``fusing'' step that minimizes the adverse effects of removing neurons. Specifically, we introduce a reweighting procedure for a neuron's connections that transfers the contributions of the pruned neurons to the ones that are retained (\S\ref{sec:reweighting}). 
\end{list}
The combination of both ideas is called \algo. We perform several experiments to validate \algo and compare to previous neuron pruning approaches, which \algo consistently outperforms. Notably, \algo's reweighting strategy benefits other pruning approaches.

\paragraph{Related work.}
Due to their large number of parameters, deep neural networks typically have a heavy memory footprint. Moreover, in many deep neural network models parameters show a significant amount of redundancy~\citep{denil-2013}. Consequently, there has been significant interest in developing techniques for reducing a network's size without penalizing its performance.  

A common approach to reducing the number of parameters is to remove connections between layers. In~\citep{lecun-1990,hassibi-1993}, connections are deleted using information drawn from the Hessian of the network's error function. \citet{sainath-2013} reduce the number of parameters by analyzing the weight matrices, and applying low-rank  factorization to the final weight layer. \citet{han-2015} remove connections with weights smaller than a given threshold before retraining the network. These methods focus on deleting parameters whose removal influences the network the least, while \algo seeks diversity and merges similar neurons; these methods can thus be used in conjunction with ours. 
Although methods such as~\citep{lecun-1990} that remove connections between layers may also delete neurons from the network by removing all of their outgoing or incoming connections, it is likely that the overall impact on the size of the network will be lesser than approaches such as \algo that remove entire neurons: indeed, removing a neuron decreases the number of rows or columns of the weight matrices connecting the neuron's layer to both the previous and following layers.

Convolutional Neural Networks~\citep{lecun-1998} replace fully-connected layers with convolution and subsampling layers, which significantly decreases the number of parameters. However, as CNNs still maintain fully-connected layers, they also benefit from \algo. 

Closer to our approach of reducing the number of hidden neurons is~\citep{he-2014}, which evaluates each hidden neuron's importance and deletes neurons with the smaller importance scores. In~\citep{DBLP:journals/corr/SrinivasB15}, a neuron is pruned when its weights are similar to those of another neuron in its layer, leading to a weight update procedure that is somewhat similar in idea (albeit simpler) to our reweighting step: where~\citep{DBLP:journals/corr/SrinivasB15} removes neurons with equal or similar weights, we consider the more complicated task of merging neurons that, as a group, perform redundant calculations based on their activations.

Other recent approaches consider network compression without pruning: in~\citep{hinton-2015}, a new, smaller network is trained on the outputs of the large network; \citet{chen-2015} use hashing to reduce the size of the weight matrices by forcing all connections within the same hash bucket to have the same weight. \citet{courbariaux-2014} and \citet{gupta-2015} show that networks can be trained and run using limited precision values to store the network parameters, thus reducing the overall memory footprint. 

We emphasize that \algo's focus on neuronal diversity is orthogonal and complementary to prior network compression techniques. Consequently, \algo can be combined, in most cases trivially, with previous approaches to memory footprint reduction. 

\section{Diversity and redundancy reduction}
In this section we introduce our technique for modeling neuronal diversity more formally. 

Let $\mathcal T$ denote the training data, $\ell$ a layer of $n_\ell$ neurons, $a_{ij}$ the activation of the $i$-th neuron on input $t_j$, and $v_i = (a_{i1},\ldots, a_{i n_\ell})^\top$ the activation vector of the $i$-th neuron obtained by feeding the training data through the network. To enforce diversity in layer $\ell$, we must determine which neurons are computing redundant information and remove them.  Doing so requires finding a maximal subset of  (linearly) independent activation vectors in a layer and retaining only the corresponding neurons. In practice, however, the number of items in the training set (or the number of batches) can be much larger than the number of neurons in a layer, so the activation vectors $v_1, \ldots, v_{n_\ell}$ are likely linearly independent. Merely selecting by the maximal subset may thus lead to a trivial solution that selects all neurons.

Reducing redundancy therefore requires a more careful approach to sampling. We propose to select a subset of neurons whose activation patterns are diverse while contributing to the network's overall computation (i.e., their activations are not saturated at 0). We achieve this diverse selection by formulating the neuron selection task as sampling from a Determinantal Point Process (DPP). We describe the details below.

\subsection{Neuronal diversity via Determinantal Point Processes}
\label{sec:pruning}
DPPs are probability measures over subsets of a ground set of items. Originally introduced to model the repulsive behavior of fermions~\citep{macchi-1975}, they have since been used fruitfully in machine-learning~\citep{kulesza-2012b}. Interestingly, they have also been recently applied to modeling inter-neuron inhibitions in neural spiking behavior in the rat hippocampus~\citep{snoek-2013}.

DPPs present an elegant mathematical technique to model diversity: the probability mass associated to each subset is proportional to the determinant (hence the name) of a DPP kernel matrix. The determinant encodes negative associations between variables, and thus DPPs tend to assign higher probability mass to diverse subsets (corresponding to diverse submatrices of the DPP kernel). Formally, a ground set of $N$ items $\Yc = \{1, \ldots, N\}$ and a probability $\Pc : 2^{\Yc} \to [0,1]$ such that
\begin{equation}
  \Pc(Y) = \frac{\det(L_Y)}{\det(L+I)},
  \label{eq:dpp}
\end{equation}
where $L$ is a $N$-by-$N$ positive definite matrix, form a DPP. $L$ is called the \emph{DPP kernel}; here, $L_Y$ indicates the $|Y| \times |Y|$ principal submatrix of $L$ indexed by the elements of $Y$. 

The key ingredient that remains to be specified is the DPP kernel, which we now describe.

\subsubsection{Constructing the DPP kernel}
There are numerous potential choices for the DPP kernel. We found that experimentally a well-tuned Gaussian RBF kernel provided a good balance between simplicity and quality: for instance, it provides much better results that simple linear kernels (obtained via the outer product of the activation vectors) and is easier to use than more complex Gaussian RBF kernels with additional parameters. A more thorough evaluation of kernel choice is future work.

Recall that layer $\ell$ has activations $v_1, \ldots, v_{n_\ell}$. Using these, we first create an $n_\ell \times n_\ell$ kernel $L'$ with bandwidth parameter $\beta$ by setting 
\begin{equation}
  L'_{ij} = \exp(-\beta \|v_i - v_j\|^2)\qquad 1 \le i, j \le n_\ell.
  \label{eq:gaussian-kernel}
\end{equation}
To ensure strict positive definiteness of the kernel matrix $L'$, we add a small diagonal perturbation $\varepsilon I$ to $L'$ ($\varepsilon = 0.01$). The choice of the bandwidth parameter could be done by cross-validation, but that would greatly increase the training cost. Therefore, we use the fixed choice $\beta = 10/|\mathcal T|$, which was experimentally seen to work well. 

Finally, in order to limit rounding errors, we introduce a final scaling operation: suppose we wish to obtain a desired size, say $k$, of sampled subsets (in which case we are said to be using a $k$-DPP~\citep{kulesza2011kdpps}). To that end, we can scale the kernel $L' + \varepsilon I$ by a factor $\gamma$, so that its \emph{expected} sample size becomes $k$. For a DPP with kernel $L$, the expected sample size is given by~\citep[Eq. 34]{kulesza-2012b}:
\[\mathbb E[|Y|] = \Tr(L (I+L)^{-1}).\]
Therefore, we scale the kernel to $\gamma(L'+\varepsilon I)$ with $\gamma$ such that
\[\gamma = \frac{k}{n_\ell-k} \cdot \frac{n_\ell-k'}{k'},\] 
where $k'$ is the expected sample size for the kernel $L' + \varepsilon I$. 

Finally, generating and then sampling from $L = \gamma (L' + \varepsilon I)$ has $\mathcal O(n_\ell^3 + n_\ell^2 |\mathcal T|)$ cost. In our experiments, this sampling cost was negligible compared with the cost of training. For networks with very large hidden layers, one can avoiding the $n_\ell^3$ cost by using more scalable sampling techniques~\citep{chengtao-2015,kang-2013}.

\subsection{Fusing redundant neurons}
\label{sec:reweighting}
Simply excising the neurons that are not sampled by the DPP drastically alters the neuron inputs to the next layer. Intuitively, since activations of neurons marked redundant are not arbitrary, throwing them away is wasteful. Ideally we should preserve the total information of a given layer, which suggests that we should ``fuse'' the information from unselected neurons into the selected ones. We achieve this via a reweighting procedure as outlined below.

Without loss of generality, let neurons 1 through $k$ be the ones sampled by the DPP and $v_1, \ldots, v_k$ their corresponding activation vectors. Let $w_{ij}$ be the weights connecting the $i$-th neuron ($1\le i\le k$) in the current layer to the $j$-th neuron in the next layer; let $\widetilde w_{ij} = \delta_{ij} + w_{ij}$ denote the updated weights after merging the contributions from the removed neurons.

We seek to minimize the impact of removing $n_\ell - k$ neurons from layer $\ell$. To that end, we minimize the difference in inputs to neurons in the subsequent layer before ($\sum_{i \leq n_\ell} w_{ij} v_i$) and after ($\sum_{i = 1 \leq k} \widetilde w_{ij}v_i$) DPP pruning. That is, we wish to solve for all neurons in the next layer (indexed by $j$, $1 \le j \le n_{\ell + 1}$):
\begin{equation}
  \label{eq:reweighting}
  \min_{\widetilde w_{ij} \in \set R}\left\| \sum\nolimits_{i = 1}^k \widetilde w_{ij}v_i - \sum\nolimits_{i=1}^{n_\ell} w_{ij} v_i\right\|_2 =  \min_{\delta_{ij} \in \set R} \left\| \sum\nolimits_{i = 1}^k \delta_{ij}v_i - \sum\nolimits_{i=k+1}^{n_\ell} w_{ij} v_i\right\|_2.
\end{equation}

Eq.~\ref{eq:reweighting} is minimized when $\sum_{i \leq k} \delta_{ij}v_i$ is the projection of $\sum_{i>k} w_{ij} v_i$ onto the linear space generated by $\{v_1, \ldots, v_k\}$. Thus, to minimize Eq.~\ref{eq:reweighting}, we obtain the coefficients $\alpha_{ij}$ that for $j > k$ minimize 
\[\left\|v_j - \sum\nolimits_{i = 1}^k \alpha_{ij} v_i\right\|_2\]

and then update the weights by setting 
\begin{equation}
  \label{eq:weight-update}
  \forall i,\, 1 \leq i \leq k, \,\widetilde w_{ij} = w_{ij} + \sum\nolimits_{r=k+1}^{n_\ell} \alpha_{ir}w_{rj}
\end{equation}
Using ordinary least squares to obtain $\alpha$, the reweighting procedure runs in $\mathcal O(|\mathcal T|n_\ell^2 + n_\ell^3)$.

\section{Experimental results}
To quantify the performance of our algorithm, we present below the results of experiments\footnote{Run in MATLAB, based on the code from DeepLearnToolBox (\url{https://github.com/rasmusbergpalm/DeepLearnToolbox}) and Alex Kulesza's code for DPPs (\url{http://web.eecs.umich.edu/~kulesza/}), on a Linux Mint system with 16GB of RAM and an i7-4710HQ CPU @ 2.50GHz.} on common datasets for neural network evaluation: \MNIST{}~\citep{mnist}, \ROT{}~\citep{larochelle-2007} and \CIFAR{}~\citep{cifar-10}. 

All networks were trained up until a certain training error threshold, using softmax activation on the output layer and sigmoids on other layers; see Table~\ref{tab:networks} for more details. In all following plots, error bars represent standard deviations.

\begin{table}[h]\small
  \caption{\small Overview of the sets of networks used in the experiments. We train each class of networks until the first iteration of backprop for which the training error reaches a predefined threshold.}
  \label{tab:networks}
  \centering
  \begin{tabular}[h]{|c|c|c|c|}\hline
     Dataset & Instances & Trained up until & Architecture \\ \hline
    \MNIST{} & 5 & $< 1\%$ error & 784 - 500 - 500 - 10 \\ \hline 
    \ROT{}   & 5 & $< 1\%$ error & 784 - 500 - 500 - 10 \\ \hline 
    \CIFAR{} & 5 & $< 50\%$ error & 3072 - 1000 - 1000 - 1000 - 10 \\ \hline 
  \end{tabular}
\end{table}

\subsection{Pruning and reweighting analysis}
\label{sec:experiments-1}
To validate our claims on the benefits of using DPPs and fusing neurons, we compare these steps separately and also simultaneously against random pruning, where a fixed number of neurons are chosen uniformly at random from a layer and then removed, with and without our fusing step. We present performance results on the test data; of course, both DPP selection and reweighting are based solely on information drawn from the training data.

\begin{figure}[h]
  \centering
  \begin{subfigure}{\linewidth}
  \centering
    \includegraphics[width=.8\linewidth]{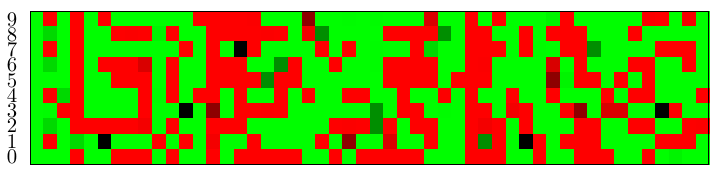}
    \caption{50 neurons sampled via DPP from the first hidden layer}
    \label{fig:reconstruction-dpp}
  \end{subfigure}
   \vskip 6pt
  \begin{subfigure}{\linewidth}
  \centering
    \includegraphics[width=.8\linewidth]{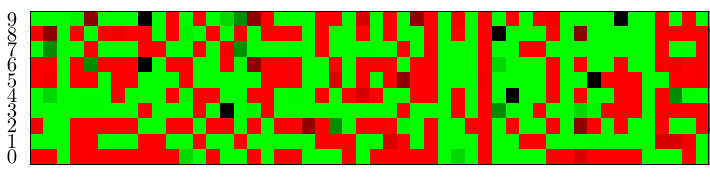}
    \caption{First 50 neurons of the first hidden layer}
    \label{fig:reconstruction-first}
  \end{subfigure}
  \caption{\small Heat map of the activation of subsets of 50 neurons for one instance of each class of the \MNIST{} dataset. The rows correspond to digits 0 through 9. Each column corresponds to the activation values of one neuron in the network's first layer on images of digits 0 through 9. Green values are activations close to 1, red values are activations close to 0.}
  \label{fig:reconstruction}
\end{figure}

Figure~\ref{fig:reconstruction} visualizes neuron activations in the first hidden layer of a network trained on the \MNIST{} dataset. Each column in the plotted heat maps represents the activation of a neuron on instances of digits 0 through 9. Figure~\ref{fig:reconstruction-dpp} shows the activations of the 50 neurons sampled using a $k$-DPP ($k$ = 50) defined over the first hidden layer, whereas Figure~\ref{fig:reconstruction-first} shows the activations of the first 50 neurons of the same layer. Figure~\ref{fig:reconstruction-first} contains multiple similar columns: for example, there are 3 entirely green columns, corresponding to three neurons that saturate to 1 on each of the 10 instances. In contrast, the DPP samples neurons with diverse activations, and Figure~\ref{fig:reconstruction-dpp} shows no similar redundancy.

\begin{figure*}[!h]
  \centering
  \begin{subfigure}{.45\linewidth}
    \includegraphics[width=\linewidth]{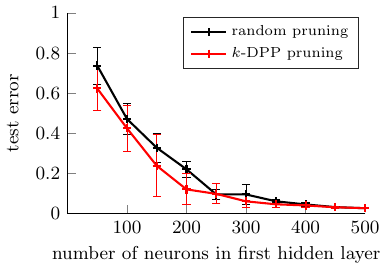}
    \label{fig:k-dpp-test-error}
    \vskip -.4cm
    \caption{\MNIST{} dataset}
  \end{subfigure}
  \begin{subfigure}{.45\linewidth}
    \includegraphics[width=\linewidth]{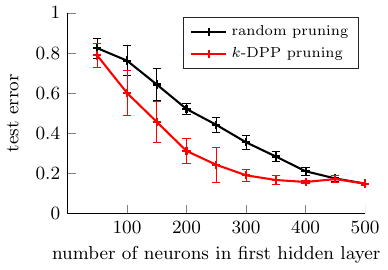}
    \label{fig:k-dpp-training-error}
    \vskip -.4cm
    \caption{\ROT{} dataset}
  \end{subfigure}
  \caption{Comparison of random and $k$-DPP pruning procedures.}
  \label{fig:k-dpp}
  \centering
  \begin{subfigure}{.45\linewidth}
    \includegraphics[width=\linewidth]{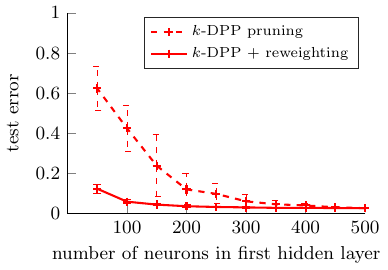}
    \label{fig:reweighting-influence-test-error}
    \vskip -.4cm
    \caption{\MNIST{} dataset}
  \end{subfigure}
  \begin{subfigure}{.45\linewidth}
    \includegraphics[width=\linewidth]{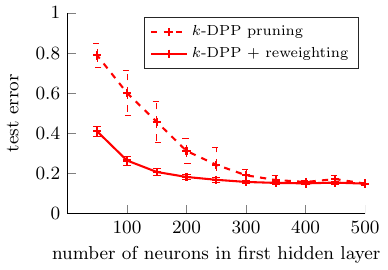}
    \label{fig:reweighting-influence-training-error}
    \vskip -.4cm
    \caption{\ROT{} dataset}
  \end{subfigure}
  \caption{Comparison of \algo ($k$-DPP + reweighting) to simple $k$-DPP pruning.}
  \label{fig:reweighting-influence}
  \centering
  \begin{subfigure}{.45\linewidth}
    \includegraphics[width=\linewidth]{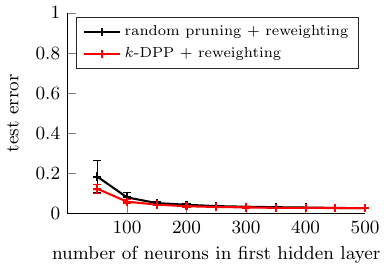}
    \label{fig:reweighting-pruning-influence-test-error}
    \vskip -.4cm
    \caption{\MNIST{} dataset}
  \end{subfigure}
  \begin{subfigure}{.45\linewidth}
    \includegraphics[width=\linewidth]{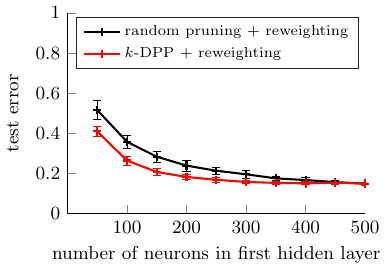}
    \label{fig:reweighting-pruning-influence-training-error}
    \vskip -.4cm
    \caption{\ROT{} dataset}
  \end{subfigure}
  \caption{Comparison of random and $k$-DPP pruning when both are followed by reweighting.}
  \label{fig:reweighting-pruning-influence}
\end{figure*}

Figures~\ref{fig:k-dpp} through~\ref{fig:reweighting-pruning-influence} illustrate the impact of each step of \algo separately. Figure~\ref{fig:k-dpp} shows the impact of pruning on test error using DPP pruning and random pruning (in which a fixed number of neurons are selected uniformly at random and removed from the network). DPP-pruned networks have consistently better training and test errors than networks pruned at random for the same final size. As expected, the more neurons are maintained, the less the error suffers from the pruning procedure; however, the pruning is in both cases destructive, and is seen to significantly increase the error rate.

This phenomenon can be mitigated by our reweighting procedure, as shown in Figure~\ref{fig:reweighting-influence}. By fusing and reweighting neurons after pruning, the training and test errors are considerably reduced, even when 90\% of the layer's neurons are removed. Pruning also reduces variability of the results: the standard deviation for the results of the reweighted networks is significantly smaller than for the non-reweighted networks, and may be thus seen as a way to regularize neural networks.

Finally, Figure~\ref{fig:reweighting-pruning-influence} illustrates the performance of \algo (DPP pruning and reweighting) compared to random pruning with reweighting. Although \algo's performance is ultimately better, the reweighting procedure also dramatically benefits the networks that were pruned randomly.

We also ran these experiments on networks for shrinking the second layer while maintaining the first layer intact. The results are similar, and may be found in Appendix~\ref{appendix:second-layer}. Notably, we found that the gap between \algo and random pruning's performances was much wider when pruning the last layer. We believe this is due to the connections to the output layer being learned much faster, thus letting a small, diverse subset of neurons (hence well suited to DPP sampling) in the last hidden layer take over the majority of the computational effort.

\subsection{Performance analysis}
\label{sec:experiments-2}
Much attention has been given to reducing the size of neural networks in order to reduce memory consumption. When using neural nets locally on devices with limited memory, it is crucial that their memory footprint be as small as possible.

\begin{figure}[t]
  \centering
  \begin{subfigure}{.45\linewidth}
    \includegraphics[width=\linewidth]{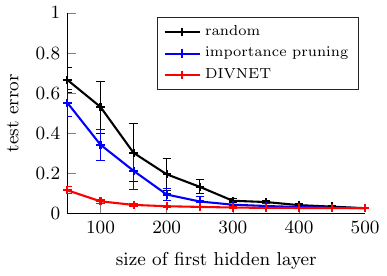}
    \label{fig:activation-pruning-mnist}
    \vskip -.4cm
    \caption{\MNIST{} dataset}
  \end{subfigure}
  \begin{subfigure}{.45\linewidth}
    \includegraphics[width=\linewidth]{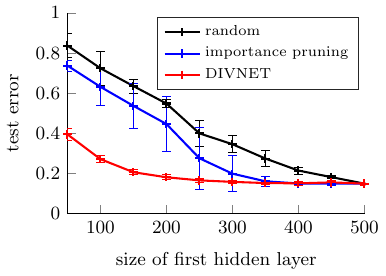}
    \label{fig:activation-pruning-rot}
    \vskip -.4cm
    \caption{\ROT{} dataset}
  \end{subfigure}
  \caption{\small Comparison of random pruning, importance pruning, and \algo's impact on the network's performance after decreasing the number of neurons in the first hidden layer of a network.}
  \label{fig:activation-pruning}
\end{figure}

Node importance-based pruning (henceforth ``importance pruning'') is one of the most intuitive ways to cut down on network size. Introduced to deep networks by~\cite{he-2014}, this method removes the neurons whose calculations impact the network the least. Among the three solutions to estimating a neuron's importance discussed in~\cite{he-2014}, the sum the output weights of each neuron (the `onorm' function) provided the best results:

\[\text{onorm}(n_i) := \frac 1 {n_{\ell+1}} \sum\nolimits_{j = 1}^{n_\ell} |w_{ij}^{\ell + 1}|.\]

Figure~\ref{fig:activation-pruning} compares the test data error of the networks after being pruned using importance pruning that uses onorm as a measure of relevance against \algo. Since importance pruning deletes neurons that contribute the least to the next layer's computations, it performs well up to a certain point; however, when pruning a significant amount of neurons, this pruning procedure even removes neurons performing essential calculations, hurting the network's performance significantly. However, since \algo  fuses redundant neurons, instead of merely deleting them its resulting network delivers much better performance even with very large amounts of pruning.

In order to illustrate numerically the impact of \algo on network performance, Table~\ref{tab:compression-train} shows network training and test errors (between 0 and 1) under various compression rates obtained with \algo, without additional retraining (that is, the pruned network is not retrained to re-optimize its weights).
\begin{table}[!h]\small
  \caption{\small Training and test error for different percentages of remaining neurons (mean $\pm$ standard deviation). Initially, \MNIST{} and \ROT{} nets have 1000 hidden neurons, and \CIFAR{} have 3000.}
  \label{tab:compression-train}
  \centering
{\setlength{\tabcolsep}{1.5mm}
  \begin{tabular}[h]{|cc|ccccc|} \hline
    \multicolumn{2}{|c|}{Remaining hidden neurons} & 10\% & 25\% & 50\% & 75\% & 100 \%\\ \hline

    \multirow{2}{*}{\MNIST{}} & training error & 0.76 $\pm$ 0.06 & 0.28 $\pm$ 0.12 & 0.15 $\pm$ 0.04 & 0.06 $\pm$ 0.04 & 0.01 $\pm 0.001$  \\ 
                                                   & test error & 0.76 $\pm$ 0.07 & 0.29 $\pm$ 0.12 & 0.17 $\pm$ 0.05 & 0.07 $\pm$ 0.03 & 0.03 $\pm$ 0.002 \\ \hline

    \multirow{2}{*}{\ROT{}} & training error & 0.74 $\pm$ 0.08 & 0.54 $\pm$ 0.09 & 0.34 $\pm$ 0.06 & 0.20 $\pm$ 0.03 & $0.01 \pm 0.003$ \\ 
       & test error & 0.73 $\pm$ 0.09 & 0.49 $\pm$ 0.11 & 0.25 $\pm$ 0.07 & 0.06 $\pm$ 0.03 & 0.15 $\pm 0.008$ \\ \hline 

    \multirow{2}{*}{\CIFAR{}} & training error & 0.84 $\pm$ 0.05 & 0.61 $\pm$ 0.01 & 0.52 $\pm$ 0.01 & 0.50 $\pm$ 0.01 & 0.49 $\pm$ 0.004 \\
     & test error & 0.85 $\pm$ 0.05 & 0.62 $\pm$ 0.02 & 0.54 $\pm$ 0.01 & 0.52 $\pm$ 0.01 & 0.51 $\pm$ 0.005 \\ \hline
  \end{tabular}}
\end{table}

\subsection{Discussion and Remarks}
\begin{list}{$\circ$}{\leftmargin=1em}
  \setlength{\itemsep}{0pt}
  \item In all experiments, sampling and reweighting ran several orders of magnitude faster than training; reweighting required significantly more time than sampling. If \algo must be further sped up, a fraction of the training set can be used instead of the entire set, at the possible cost of subsequent network performance.
  \item When using DPPs with a Gaussian RBF kernel, sampled neurons need not have linearly independent activation vectors: not only is the DPP sampling probabilistic, the kernel itself is not scale invariant. Indeed, for two collinear but unequal activation vectors, the corresponding coefficient in the kernel will not be 1 (or $\gamma$ with the $L \leftarrow \gamma L$ update).
  \item In our work, we selected a subset of neurons by sampling once from the DPP. Alternatively, one could sample a fixed amount of times, using the subset with the highest likelihood (i.e., largest $\det(L_Y)$), or greedily approximate the mode of the DPP distribution.
  \item Our reweighting procedure benefits competing pruning methods as well (see Figure~\ref{fig:reweighting-pruning-influence}).
  \item We also investigated DPP sampling for pruning concurrently with training iterations, hoping that this might allow us to detect superfluous neurons before convergence, and thus reduce training time. However, we observed that in this case DPP pruning, with or without reweighting, did not offer a significant advantage over random pruning.
  \item Consistently over all datasets and networks, the expected sample size from the kernel $L'$ was much smaller for the last hidden layer than for other layers. We hypothesize that this is caused by the connections to the output layer being learned faster than the others, allowing a small subset of neurons to take over the majority of the computational effort.
\end{list}

\section{Future work and conclusion}
\algo leverages similarities between the behaviors of neurons in a layer to detect redundant parameters and merge them, thereby enforcing neuronal diversity within each hidden layer. Using \algo, large, redundant networks can be shrunk to much smaller structures without impacting their performance and without requiring further training. We believe that the performance profile of \algo will remain similar even when scaling to larger scale datasets and networks, and hope to include results on bigger problems (e.g., Imagenet~\citep{ILSVRC15}) in the future.

Many hyper-parameters can be tuned by a user as per need include: the number of remaining neurons per layer can be fixed manually; the precision of the reweighting and the sampling procedure can be tuned by choosing how many training instances are used to generate the DPP kernel and the reweighting coefficients, creating a trade-off between accuracy, memory management, and computational time. Although \algo requires the user to select the size of the final network, we believe that a method where no parameter explicitly needs to be tuned is worth investigating. The fact that DPPs can be augmented to also reflect different distributions over the sampled set sizes~\citep[\textsection 5.1.1]{kulesza-2012b} might be leveraged to remove the burden of choosing the layer's size from the user.

Importantly, \algo is agnostic to most parameters of the network, as it only requires knowledge of the activation vectors. Consequently, \algo can be easily used jointly with other pruning/memory management methods to reduce size.  Further, the reweighting procedure is agnostic to how the pruning is done, as shown in our experiments.

Furthermore, the principles behind \algo can theoretically also be leveraged in non fully-connected settings. For example, the same diversifying approach may also be applicable to filters in CNNs: if a layer of the CNN is connected to a simple, feed-forward layer -- such as the S4 layer in~\cite{lecun-1998} -- by viewing each filter's activation values as an vector and applying \algo on the resulting activation matrix, one may be able to remove entire filters from the network, thus significantly reducing CNN's memory footprint.

Finally, we believe that investigating DPP pruning with different kernels, such as kernels invariant to the scaling of the activation vectors, or even kernels learned from data, may provide insight into which interactions between neurons of a layer contain the information necessary for obtaining good representations and accurate classification. This marks an interesting line of future investigation, both for training and representation learning. 

\subsubsection*{Acknowledgments}
This work is partly supported by NSF grant: IIS-1409802.

\bibliographystyle{abbrvnat}
\setlength{\bibsep}{.33pt}
{\footnotesize \bibliography{submission}}

\clearpage
\appendix
\section{Pruning the second layer}
\label{appendix:second-layer}

\begin{figure*}[!h]
  \centering
  \begin{subfigure}{.49\linewidth}
    \includegraphics[width=\linewidth]{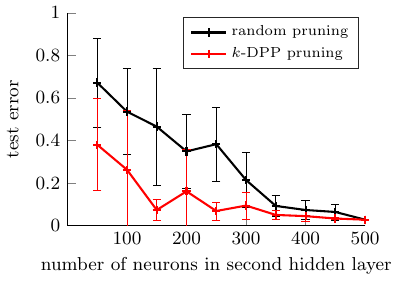}
    \caption{\MNIST{} dataset}
  \end{subfigure}
  \begin{subfigure}{.49\linewidth}
    \includegraphics[width=\linewidth]{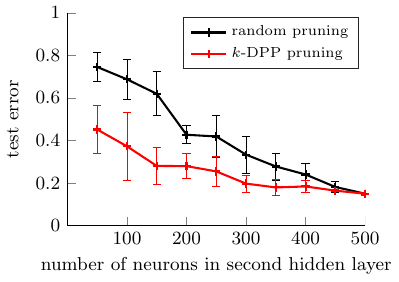}
    \caption{\ROT{} dataset}
  \end{subfigure}
  \caption{Comparison of random and $k$-DPP pruning procedures.}
\end{figure*}

\begin{figure*}[!h]
  \centering
  \begin{subfigure}{.49\linewidth}
    \includegraphics[width=\linewidth]{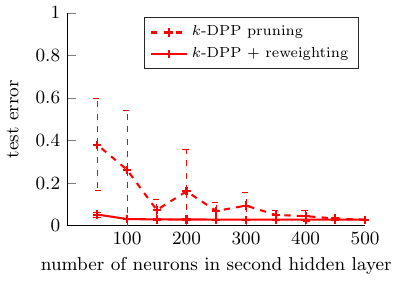}
    \caption{\MNIST{} dataset}
  \end{subfigure}
  \begin{subfigure}{.49\linewidth}
    \includegraphics[width=\linewidth]{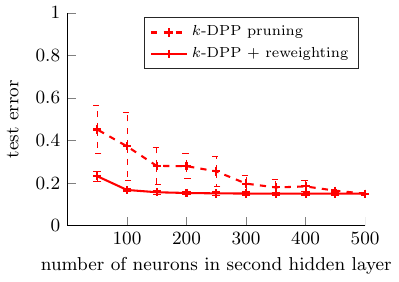}
    \caption{\ROT{} dataset}
  \end{subfigure}
  \caption{Comparison of \algo to simple $k$-DPP pruning.}
\end{figure*}

\begin{figure*}[!h]
  \centering
  \begin{subfigure}{.49\linewidth}
    \includegraphics[width=\linewidth]{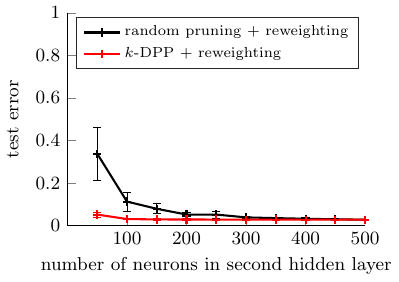}
    \caption{\MNIST{} dataset}
  \end{subfigure}
  \begin{subfigure}{.49\linewidth}
    \includegraphics[width=\linewidth]{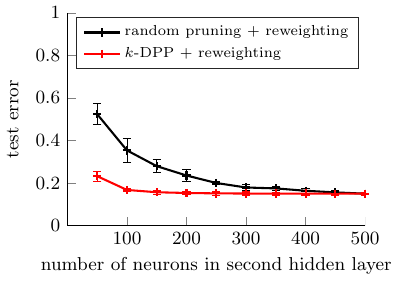}
    \caption{\ROT{} dataset}
  \end{subfigure}
  \caption{Comparison of random and $k$-DPP pruning when both are followed by reweighting.}
\end{figure*}

\clearpage
\section{Influence of the $\beta$ parameter on network size and error}
\begin{figure}[!h]
  \centering
  \includegraphics[width=.75\linewidth]{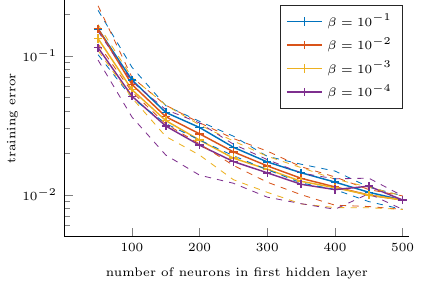}
  \caption{Influence of $\beta$ on training error (using the networks trained on \MNIST{}). The dotted lines show min and max errors.}
\end{figure}

\begin{figure}[!h]
\centering
  \includegraphics[width=.7\linewidth]{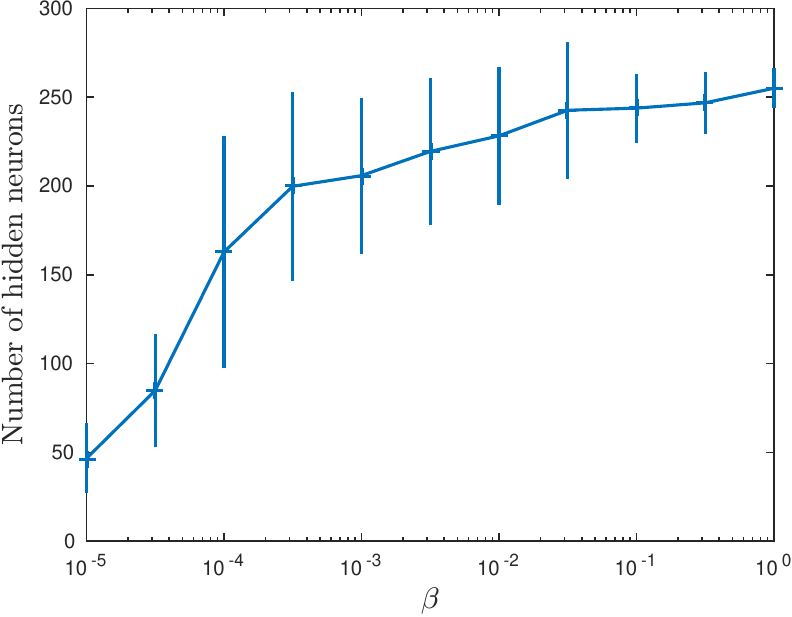}
  \caption{Influence of $\beta$ on the number of neurons that remain after pruning networks trained on \MNIST{} (when pruning non-parametrically, using a DPP instead of a $k$-DPP.)}
\end{figure}

\clearpage
\section{Comparison of \algo to importance-based pruning and random pruning on the CIFAR-10 dataset}

\begin{figure}[!h]
  \centering
  \begin{subfigure}{.49\linewidth}
    \includegraphics[width=\linewidth]{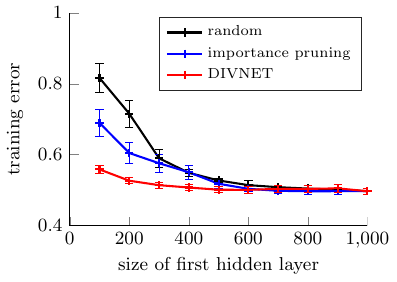}
    \caption{Training error on \CIFAR{} dataset}
  \end{subfigure}
  \begin{subfigure}{.49\linewidth}
    \includegraphics[width=\linewidth]{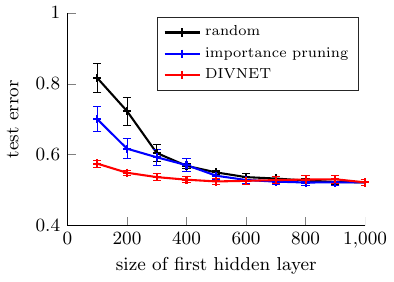}
    \caption{Test error on \CIFAR{} dataset}
  \end{subfigure}
  \caption{Comparison of random pruning, importance pruning, and \algo's impact on the network's performance after decreasing the number of parameters in the network.}
\end{figure}
\end{document}